\documentclass[letterpaper, 10 pt, conference]{ieeeconf}  

\usepackage{url}
\usepackage{graphicx}
\usepackage{booktabs}
\usepackage{romannum}
\usepackage{amsfonts}
\usepackage{multirow}

\usepackage{array}
\usepackage{makecell}
\usepackage{float}
\usepackage[export]{adjustbox} 
\usepackage{algorithm}
\usepackage{balance}
\usepackage{cite}
\usepackage{algpseudocode} 
\usepackage{longtable}
\usepackage{threeparttable}

\usepackage{booktabs}
\usepackage{graphics}
\usepackage{upgreek}
\usepackage{amssymb}
\usepackage{refstyle}

\usepackage{bbding}

\usepackage{amsmath}
\usepackage{newtxtext,newtxmath}
\usepackage{color}
\usepackage[table,dvipsnames]{xcolor}

\usepackage{subfigure}
\usepackage{mathtools}

\definecolor{c1_red}{HTML}{dc3023}
\definecolor{c1_green}{HTML}{0c8918}
\definecolor{light-blue}{HTML}{E6EEF9}

\usepackage[pagebackref=true,breaklinks=true,colorlinks=true,bookmarks=true]{hyperref}
\hypersetup{colorlinks=true,
            linkcolor=blue,
            filecolor=magenta,      
            citecolor=blue
            }

\usepackage{cleveref}

\IEEEoverridecommandlockouts                              

\title{\LARGE \bf
ContactDexNet:
Multi-fingered Robotic Hand Grasping in Cluttered Environments through Hand-Object Contact Semantic Mapping}
\author{Lei Zhang$^{1,2}$, Kaixin Bai$^{1,2\dag}$, Guowen Huang$^{2,3}$, Zhenshan Bing$^{3}$, \\ Zhaopeng Chen$^{2}$,  Alois Knoll$^{3}$, Jianwei Zhang$^{1}$
\thanks{\dag Corresponding author. kaixin.bai@studium.uni-hamburg.de }
\thanks{{$^{1}$TAMS (Technical Aspects of Multimodal Systems), Department of
Informatics, Universit\"at Hamburg, Hamburg, Germany}, {$^{2}$Agile Robots AG, Munich, Germany}, {$^{3}$Technical University of Munich, Munich, Germany}.}
}
\DeclarePairedDelimiterX{\norm}[1]{\lVert}{\rVert}{#1}

\begin{document}

\maketitle

\thispagestyle{empty}
\pagestyle{empty}

\begin{abstract}    

The deep learning models has significantly advanced dexterous manipulation techniques for multi-fingered hand grasping. However, the contact information-guided grasping in cluttered environments remains largely underexplored. To address this gap, we have developed ContactDexNet, a method for generating multi-fingered hand grasp samples in cluttered settings through contact semantic map. 
We introduce a contact semantic conditional variational autoencoder network (CoSe-CVAE) for creating comprehensive contact semantic map from object point cloud. We utilize grasp detection method to estimate hand grasp poses from the contact semantic map. Finally, an unified grasp evaluation model PointNetGPD++ is designed to assess grasp quality and collision probability, substantially improving the reliability of identifying optimal grasps in cluttered scenarios. 
Our grasp generation method has demonstrated remarkable success, outperforming state-of-the-art (SOTA) methods by at least 4.7\%, with 81.0\% average grasping success rate in real-world single-object grasping using a known hand, and by at least 9.0\% when using an unknown hand. Moreover, in cluttered scenes, our method attains a 76.7\% success rate, outperforming the SOTA method by 6.3\%. We also proposed the multi-modal multi-fingered grasping dataset generation method. Our multi-fingered hand grasping dataset outperforms previous datasets in scene diversity, modality diversity. 
More details and supplementary materials can be found at \href{https://sites.google.com/view/contact-dexnet}{https://sites.google.com/view/contact-dexnet}.
\end{abstract}

\section{Introduction}
\label{intro}

Recent advancements in multi-fingered robotic grasping research~\cite{shao2020unigrasp,li2023gendexgrasp} and human grasp generation~\cite{liu2023contactgen,jiang2021hand,yang2021cpf} have focused on leveraging hand-object contact information to guide the generation of grasping strategies. 
Specifically, contact information such as contact points from UniGrasp~\cite{shao2020unigrasp} and contact distance map from GenDexGrasp~\cite{li2023gendexgrasp} has been shown to enhance the generalizability of grasp generation for previously unknown robotic hands. Additionally, contact maps can facilitate the synthesis of functional human grasp postures~\cite{wu2023functional}. However, existing approaches still face challenges in contact information-guided robotic grasp generation, including low robustness due to sparse contact points~\cite{shao2020unigrasp}, semantic ambiguity arising from the absence of contact semantic information~\cite{li2023gendexgrasp}, high model complexity~\cite{liu2023contactgen}, and infeasible grasp poses resulting from the structural differences between human and robotic hands~\cite{liu2023contactgen}. 
To address these limitations, we propose the CoSe-CVAE to generate contact semantic maps to improve multi-fingered robotic hand grasp generation.

\begin{figure}[htbp]
    \begin{center}

        \includegraphics[width=8cm]{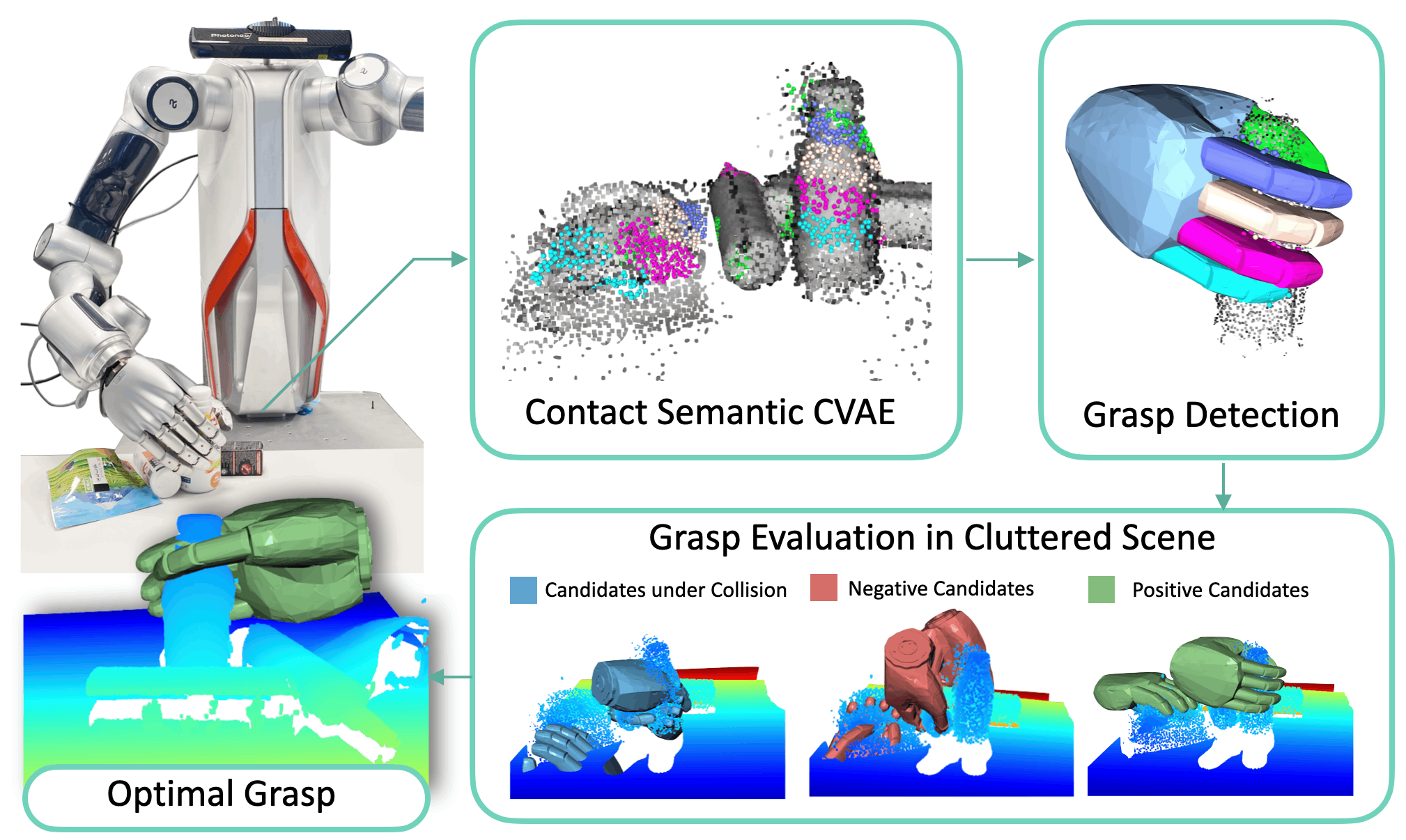}
		\caption{ 
  Employing CoSe-CVAE, the contact semantic maps are derived from object point clouds. Grasp detection leverages contact prior information for estimating grasp poses. Subsequent grasp evaluation model PointNetGPD++ assesses both grasping qualities with collision awareness to identify the optimal grasp in cluttered settings. (Blue: grasp candidates colliding with the surroundings, Red: negative grasp candidates, Green: Positive grasp candidates. )}
		\label{fig.head_photo}
	\end{center}
\end{figure}

Real-world robotic grasping scenarios are often cluttered, making grasp planning for multi-fingered hands particularly challenging. 
While numerous studies have explored grasping in such scenarios using two-jaw grippers~\cite{sundermeyer2021contact,breyer2021volumetric,wang2021graspness} and multi-fingered robotic hands~\cite{duan2024learning,li2022hgc,li2023planning,corsaro2021learning,wu2019pixel}, existing multi-fingered robotic grasping methods still struggle to achieve robust and reliable performance. 
Specifically, many approaches fail to effectively evaluate and execute accurate multi-fingered grasps when potential collisions with surrounding objects exist~\cite{li2022hgc}. Grasp prediction errors may cause premature finger contact, pushing the object away instead of achieving a stable grasp. This issue can be mitigated by incorporating contact information between perception and grasp execution. 
Furthermore, many existing methods are designed for specific robotic hand models, limiting their adaptability across different multi-fingered hands~\cite{li2022hgc}. 
This lack of generalizability increases the cost of data collection and model training for grasping with unknown robotic hands. 
Inspired by our previous work~\cite{liang2019pointnetgpd}, we introduce a generalizable grasp evaluator PointNetGPD++ that estimates grasp quality and collision probability. Designed for broad applicability, our approach enhances adaptability across different robotic hand models and diverse grasping scenarios.

\begin{table*}[]
    \centering
    \scriptsize
    \vspace*{5mm}
        \caption{Comparison of Multi-fingered Robotic Hand Grasping Datasets.}
        \begin{tabular}{cccccccc}

    \hline
    Methods & Hand Type & Cluttered Scene & Grasp Quality & Evaluation Metric & Contact Distance & Contact Semantic &  Affordance  \\
    \hline
    ContactPose~\cite{brahmbhatt2020contactpose}, GRAB~\cite{taheri2020grab} & Human & {\textcolor{c1_red}{\XSolidBrush}} & {\textcolor{c1_red}{\XSolidBrush}} & - & {\textcolor{c1_green}{\checkmark}} & {\textcolor{c1_green}{\checkmark}} & {\textcolor{c1_red}{\XSolidBrush}} \\
    DexYCB~\cite{chao2021dexycb} & Human & {\textcolor{c1_green}{\checkmark}} & {\textcolor{c1_red}{\XSolidBrush}} & - & {\textcolor{c1_green}{\checkmark}} & {\textcolor{c1_red}{\XSolidBrush}} & {\textcolor{c1_red}{\XSolidBrush}} \\
    GanHand~\cite{corona2020ganhand} & Human & {\textcolor{c1_green}{\checkmark}} & {\textcolor{c1_red}{\XSolidBrush}} & - & {\textcolor{c1_green}{\checkmark}} & {\textcolor{c1_red}{\XSolidBrush}} & {\textcolor{c1_green}{\checkmark}} \\
    DDGC~\cite{lundell2021ddgc} & Robot & {\textcolor{c1_green}{\checkmark}} & {\textcolor{c1_green}{\checkmark}} & GraspIt!~\cite{miller2004graspit} & {\textcolor{c1_red}{\XSolidBrush}} & {\textcolor{c1_red}{\XSolidBrush}} & {\textcolor{c1_red}{\XSolidBrush}} \\
    Columbia Grasp Database~\cite{goldfeder2009columbia} & Robot & {\textcolor{c1_red}{\XSolidBrush}} & {\textcolor{c1_green}{\checkmark}} & GraspIt!~\cite{miller2004graspit} & {\textcolor{c1_red}{\XSolidBrush}} & {\textcolor{c1_red}{\XSolidBrush}} & {\textcolor{c1_red}{\XSolidBrush}} \\
    Fast-Grasp'D~\cite{turpin2023fast} & Robot & {\textcolor{c1_red}{\XSolidBrush}} & {\textcolor{c1_green}{\checkmark}} & Trial-and-Error & {\textcolor{c1_red}{\XSolidBrush}} & {\textcolor{c1_red}{\XSolidBrush}} & {\textcolor{c1_red}{\XSolidBrush}} \\
    DexGraspNet~\cite{wang2023dexgraspnet} & Human\&Robot & {\textcolor{c1_red}{\XSolidBrush}} & {\textcolor{c1_green}{\checkmark}} & Trial-and-Error & {\textcolor{c1_green}{\checkmark}} & {\textcolor{c1_red}{\XSolidBrush}} & {\textcolor{c1_red}{\XSolidBrush}} \\
    DexGraspNet 2.0~\cite{zhang2024dexgraspnet} & Robot & {\textcolor{c1_green}{\checkmark}} & {\textcolor{c1_green}{\checkmark}} & Trial-and-Error & {\textcolor{c1_green}{\checkmark}} & {\textcolor{c1_red}{\XSolidBrush}} & {\textcolor{c1_red}{\XSolidBrush}} \\
    GenDexGrasp~\cite{li2023gendexgrasp} & Robot & {\textcolor{c1_red}{\XSolidBrush}} & {\textcolor{c1_green}{\checkmark}} & Trial-and-Error & {\textcolor{c1_green}{\checkmark}} & {\textcolor{c1_red}{\XSolidBrush}} & {\textcolor{c1_red}{\XSolidBrush}} \\
    	
\rowcolor{light-blue}
    Ours & Robot & {\textcolor{c1_green}{\checkmark}} & {\textcolor{c1_green}{\checkmark}} & Trial-and-Error & {\textcolor{c1_green}{\checkmark}} & {\textcolor{c1_green}{\checkmark}} & {\textcolor{c1_green}{\checkmark}} \\
    \hline
    \end{tabular}

    \label{tab:multi_finger_hand_grasping_dataset}
\end{table*}

Our research introduces a contact information-guided multi-fingered robotic grasp generation pipeline in cluttered scenes, leveraging contact semantic maps to enhance grasp quality and adaptability. 
This pipeline includes the CoSe-CVAE, a grasp detection method and a generalizable grasp evaluation model (PointNetGPD++). CoSe-CVAE is designed to generate contact semantic maps, where the semantic information indicates which fingers are in contact with the object. 
Furthermore, the grasp poses are estimated based on predicted contact semantic maps and optimal grasp is selected based on grasp qualities from grasp evaluation model. Our main contributions are as follows:
\begin{enumerate}

    \item We propose a \textbf{contact semantic conditional variational autoencoder network (CoSe-CVAE)} that generates multi-fingered grasping contact semantic maps from object point clouds. CoSe-CVAE generates richer, more diverse contact point maps with semantic information, enabling more stable and reliable grasp generation guided by contact information. It improves the grasping success rate using known and unknown hands by at least 4.7\% and 9.0\%.
    
    \item We introduce a \textbf{generalizable grasp evaluation network (PointNetGPD++)} estimating grasp scores by analysing the partial scene point cloud and hand geometric features based on PointNet++~\cite{qi2017pointnet++}. 
    The network is capable of evaluating grasping in cluttered scenes for both known and unknown multi-fingered hands. Our method outperforms SOTA approaches~\cite{li2022hgc,li2023gendexgrasp,shao2020unigrasp,liu2023contactgen} in average grasp success rate by at least 4.7\% for grasping from single-object scenes and by 6.3\% for grasping from cluttered scenes.

    \item We integrate a pipeline for generating a multi-modal multi-fingered grasping dataset in cluttered environments, based on DexGraspNet~\cite{wang2023dexgraspnet}. 
     Compared to previous multi-fingered hand datasets, our dataset includes more complex scenes, a greater number of modalities, and newly introduced contact semantic maps, which enhance grasp representation. Moreover, these maps improve transferability across different robotic hands, enabling broader applicability. 
\end{enumerate}

\section{Related Work}
\label{sec:relatedwork}
\subsection{Multi-fingered Robotic Hand Grasping in Cluttered Environments}
Grasping in cluttered environments using multi-fingered robotic hands presents a significant challenge due to their high degrees of freedom and the complex collision dynamics with surrounding objects.
Although there has been extensive research on grasping in cluttered environments with two-jaw gripper~\cite{zhang2024collision,sundermeyer2021contact} and multi-fingered hand grasping from single-object scenes~\cite{lum2024dextrah,weng2024dexdiffuser,wu2024unidexfpm,duan2022learning,liu2020deep,lundell2021multi}, studies on multi-fingered robotic hand grasping in such environments remain limited~\cite{duan2024learning,li2022hgc,wu2019pixel,lundell2021ddgc,corsaro2021learning,berenson2008grasp}. 
Currently, datasets for multi-fingered robotic hand grasping in cluttered environments are severely limited. We provide an overview of existing multi-fingered hand grasping datasets and their available modalities, as summarized in Tab.~\ref{tab:multi_finger_hand_grasping_dataset}. However, there are no dataset that includes cluttered scenes while capturing all relevant multi-modal information. To date, no studies have utilized contact information to guide grasp generation in cluttered environments with multi-fingered robotic hands, and no corresponding datasets have been developed. To address this gap, we extended the existing grasp generation pipeline~\cite{wang2023dexgraspnet} to cluttered environments, producing contact semantic maps. 

\subsection{Contact Information-guided Grasping Generation}

Hand-object representations are widely used in various domains: they are crucial for generating plausible hand poses~\cite{jiang2021hand,yang2021cpf,wu2022learning}, formulating generalized representations for diverse end-effectors~\cite{shao2020unigrasp,li2023gendexgrasp}, and bridging the gap between human and robotic hand representations ~\cite{zhu2021toward,amor2012generalization}. 
Various types of contact representations are employed, including contact touch code~\cite{zhu2021toward}, contact distance or points~\cite{liu2023dexrepnet,brahmbhatt2019contactgrasp,shao2020unigrasp}, contact semantic map~\cite{liu2023contactgen,brahmbhatt2020contactpose}. 
UniGrasp~\cite{shao2020unigrasp} introduced a generalized model that sequentially generates contact points. 
Generative models, renowned for their diversity and generative capabilities, have been increasingly applied in the field of grasp generation~\cite{li2023gendexgrasp,wei2022dvgg,mousavian20196,urain2022se}. 
GenDexGrasp~\cite{li2023gendexgrasp} employed a generative model to generate contact distance maps from object point clouds. 
However, we found that grasps generated using contact distance maps lacked stability due to the absence of semantic information. 
To address these limitations, we propose a novel generative model, named CoSe-CVAE, which generates contact semantic maps from object point clouds and incorporates the grasp generation pipeline for cluttered environments.

\subsection{Grasping Evaluation in Cluttered Scenes}

Concerning collision-free grasp detection in cluttered environments, extensive research has been conducted on employing neural networks to predict collision-free grasp samples from visual data, particularly in the context of two-fingered grasping setups~\cite{zhang2024collision,li2023planning,li2022hgc,wei2021gpr,sundermeyer2021contact,breyer2021volumetric,wang2021graspness}. 
However, multi-fingered hands, with their additional joints, pose greater challenges in learning implicit collision representations. 
Previous evaluation methods~\cite{li2022hgc,mayer2022ffhnet} were often designed for a specific robotic hand and lacked the ability to generalize across different hand types. To address this complexity and identify optimal grasp candidates in cluttered environments, we develop an unified grasp evaluation model that estimate grasp scores with collision awareness.

\begin{figure}[htbp]
    \begin{center}
		\includegraphics[width=6cm]{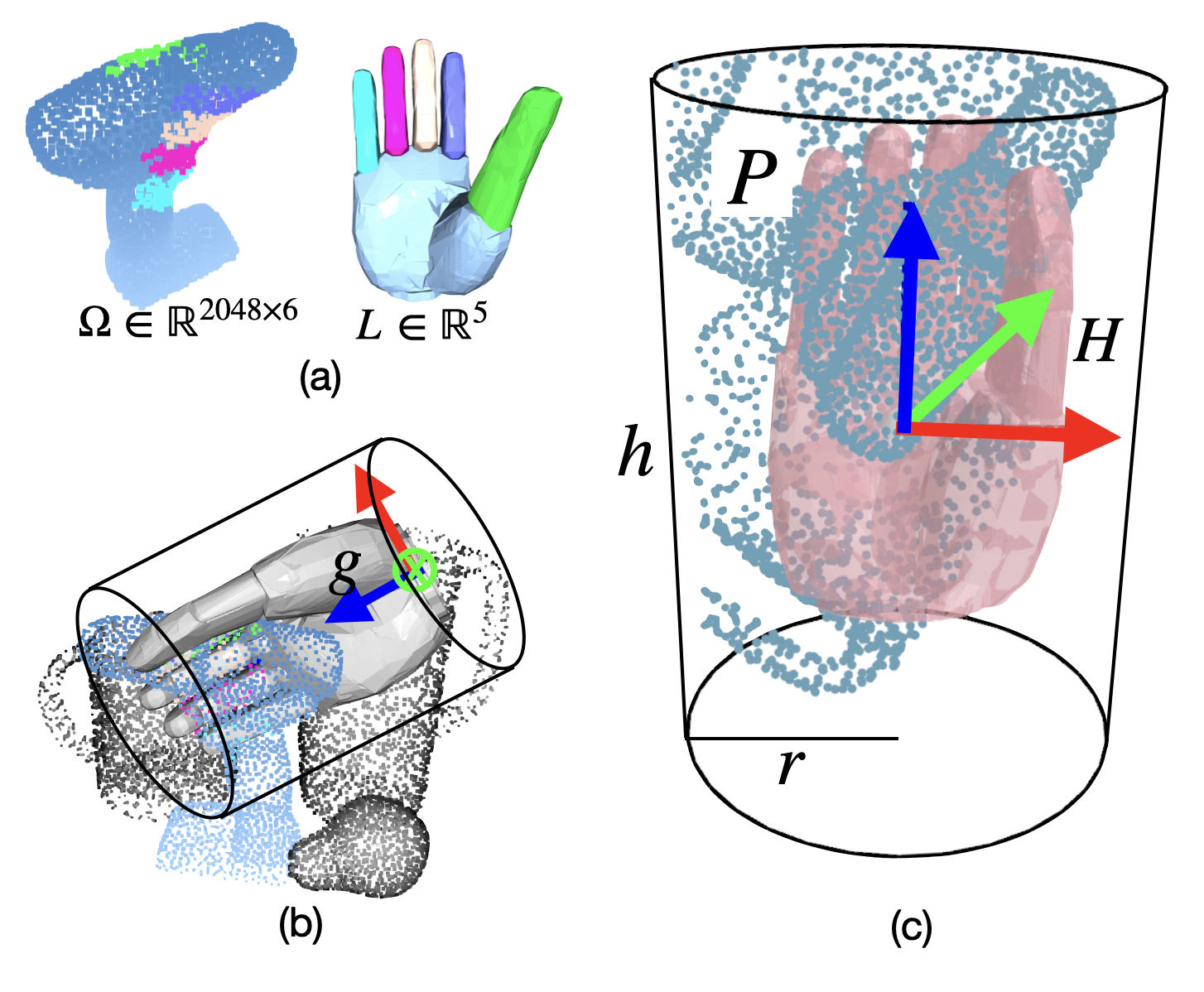}
		\caption{(a) The CoSe-CVAE model predicts contact semantic maps based on the object’s point cloud, representing both the geometric and semantic information of contact points across different fingers. (b) Grasp candidate in cluttered scene. (c) The grasp evaluation network assesses grasp quality by utilizing partial scene point cloud surrounding the grasp sample $P$, along with the sampled point cloud of the multi-fingered hand $H$.}

		\label{fig.contact_and_grasp_representation}
	\end{center}
\end{figure}

\section{Problem Statement and Methods}\label{sec:method}
\subsection{Problem Statement}

In multi-fingered robotic hand grasping tasks within cluttered scenes, it's crucial to consider both hand grasp quality and the collision probability with the surrounding unstructured environment. 

We define a robotic hand pose $g=[T, \Theta]$. 
$T$ denotes hand wrist pose, $\Theta$ represents joint poses $(\theta_{1},\theta_{1},\dots,\theta_{d})$. $d$ denotes the number of degrees of freedom (DOF), corresponding to $15$ DOF and $20$ joints in the DLR-HIT II hand~\cite{chen2010experimental}. The dataset generation of multi-fingered robotic hand grasping is detailed in Sec.~\ref{sec:dataset_generation}.

We utilize contact semantic map $\Omega \in \mathbb{R}^{2048 \times (n +1)}$ with $2048$ points to represent the contact points between $n$ fingers of robotic hand and grasped object and points without contacts, as shown in Fig.~\ref{fig.contact_and_grasp_representation} (a). 
Generative model CoSe-CVAE $f$ is able to estimate $N$ contact semantic maps from object point cloud $O$, as introduced in Sec.~\ref{sec:contact_semantic_cvae}. 
Grasp detection $F$ is presented in Sec.~\ref{sec:contact_cvae_and_grasp_detection} to estimate grasp candidates based on contact semantic maps. For estimating the optimal grasp candidate $g_{\rm optimal}$ from a cluttered scene's point cloud, grasp evaluation network $\Psi$ infers grasp qualities $q$ from partial scene point cloud $P$ and sampled hand point cloud $H$, as shown in Fig.~\ref{fig.contact_and_grasp_representation} (b) and described in Sec.~\ref{sec:grasp_evaluation_network}. 
The partial scene point cloud is obtained through filtering the original scene points using a cylindrical region in the robotic hand's frame, defined by a radius $r$ and height $h$, as depicted in Fig.~\ref{fig.contact_and_grasp_representation} (c).  
The optimal grasp pose $g_{\rm optimal}$ is selected based on these inferences. The pipeline is summarized in Eq.~\ref{eq.whole_pipeline}.

\begin{equation}
\label{eq.whole_pipeline}
\begin{split}
\bigcup_{i=0}^{N-1}  g_{i} &= \bigcup_{i=0}^{N-1}  F (f_{i} (O))\\
\bigcup_{i=0}^{N -1} q_{i} &= \bigcup_{i=0}^{N  -1} \Psi (P_{i}, H_{i})\\
g_{\rm optimal}=&\operatorname*{arg\,max}_g ~\bigcup_{i=0}^{N-1} q_i 
\end{split}
\end{equation}

\begin{figure}[htbp]
    \begin{center}

  \includegraphics[width=7cm]{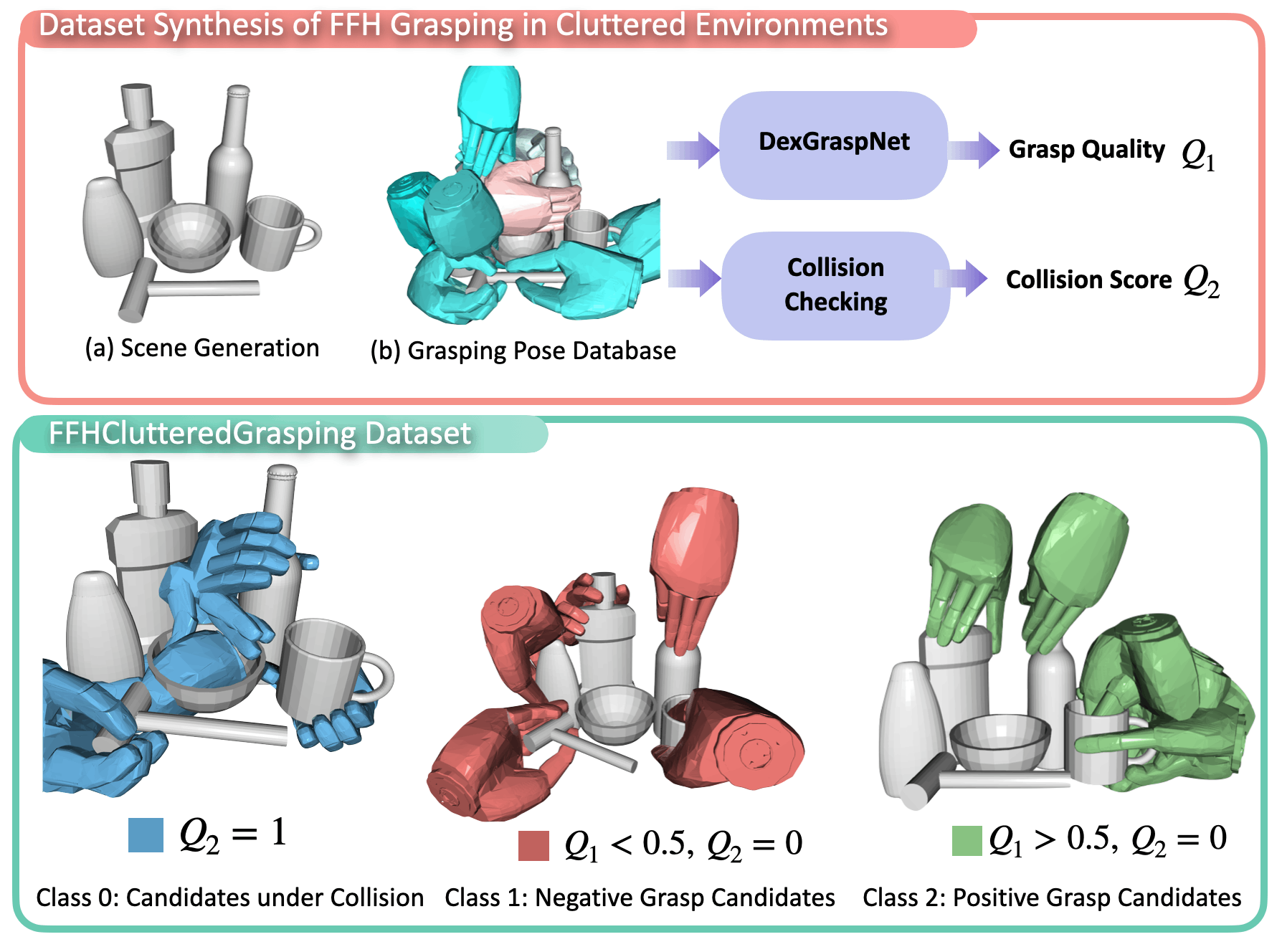}
		\caption{Pipeline for generating multi-fingered robotic hand grasps in cluttered settings, involving scene generation, hand pose generation, collision checking, grasp quality validation, and dataset labeling with grasp quality and collision status. Grasping candidates under collision condition are plotted in blue. Unreliable grasp candidates, where $Q_1 < 0.5$, are highlighted in red, while reliable grasp candidates are marked in green.}
		\label{fig.pipeline_dataset_generation}
	\end{center}
\end{figure}

\subsection{Multi-Modal Multi-fingered Hand Grasping Dataset Generation}
\label{sec:dataset_generation}

To effectively address the complexities of multi-fingered robotic hand planning in intricate environments, we have developed a method for grasping synthesis. The algorithm pipeline is summarized in Alg.~\ref{alg.grasp_generate} and shown in Fig.~\ref{fig.pipeline_dataset_generation}.


\begin{algorithm}
	\caption{Dataset Synthesis Algorithm}
	\label{alg.grasp_generate}
	\begin{algorithmic}[1]
		\State \textit{\textbf{Input}}: Object database $A$, Number of sampling grasp poses $M$,  Number of objects in scene $m$
            \State \textit{\textbf{Output}}: Set of multi-fingered robotic hand grasping candidates with grasp pose~$g$, grasp quality~$Q_{1}$, collision score~$Q_{2}$, contact semantic map~$\Omega$, contact distance map~$\Omega_d$
            \State // Generate dataset for single-object scenes.
            \For{each object in $A$ }
                \State Estimate $g$, $Q_1$ based on~\cite{wang2023dexgraspnet}, $\Omega_d$ based on~\cite{li2023gendexgrasp}.
                \State Estimate proposed $\Omega$ as detailed in~\ref{subsubsection.contact_semantic_map_estimation}.
            \EndFor
            \State // Generate dataset for cluttered scenes.
            \For{each cluttered scene}
                \State Sample $m$ objects from $A$.
                \State Construct a cluttered scene by iteratively adding objects in sampled poses. Ensure that each newly placed object does not collide with existing objects using collision detection~\cite{sundermeyer2021contact}.
                \For{each object in the generated cluttered scene}
                    \For{each each grasp candidate of the object}
                        \State Compute the collision score $Q_{2}$ between mesh of robotic hand at the candidate pose and surrounding objects using collision detection.
                        \State Obtain $(g, Q_1, Q_2, \Omega, \Omega_d)$.
                    \EndFor
                \EndFor
            \EndFor
	\end{algorithmic}  
\end{algorithm}
		

\setlength{\textfloatsep}{5pt}

\subsubsection{Contact Semantic Map Estimation}
\label{subsubsection.contact_semantic_map_estimation}
The contact semantic map is computed by estimating the nearest points on object surface to hand's fingers. Coarse-estimated nearest points on object's surface $P_n$ is calculated based on aligned distance $\epsilon(\phi_{\rm finger}, O)$~\cite{li2023gendexgrasp}, normal vector $n_o$ of object surface point and robotic hand surface point $v_h$, as formalized in Eq.~\ref{eq.contact-points}. Each finger's surface point is denoted by $\phi_{\rm finger}$ and the object point cloud is represented using $O$. 
The object point clouds with contact semantic labels are denoted by $P_{\rm contact}$.  
This process is formalized as follows:
\begin{equation}
\label{eq.contact-points}
\begin{aligned}
        P_n = & \left\{v_h - \epsilon_{min} n_{o},\forall v_h \in \phi_{\rm finger}\right\}\\
        P_{\rm contact}  = &\left\{(p^{'} ,L) \mid p^{'}\in O, 
        \exists p \in P_n  \right. \\
        &\left.\text{s.t.}  \left\|p-p^{'}\right\|<\tau~\text{and}~L = l \right\}
\end{aligned}    
\end{equation}
where, $\epsilon_{min}$ denotes the aligned distance characterized by the smallest absolute value. $\tau$ signifies the threshold parameter. The semantic label $L$ is denoted by the classification index of the fingers $l$, shown in Fig.~\ref{fig.contact_and_grasp_representation} and is used to label the contact semantic categories of the object's point cloud. 
Consequently, contact semantic map $\Omega \in \mathbb{R}^{2048 \times (n +1)}$ is shown in Fig.~\ref{fig.contact_and_grasp_representation} (a).

\subsection{Contact Semantic CVAE}

\label{sec:contact_semantic_cvae}

\begin{figure}[htbp]
    \begin{center}
		\includegraphics[width=8.5cm]{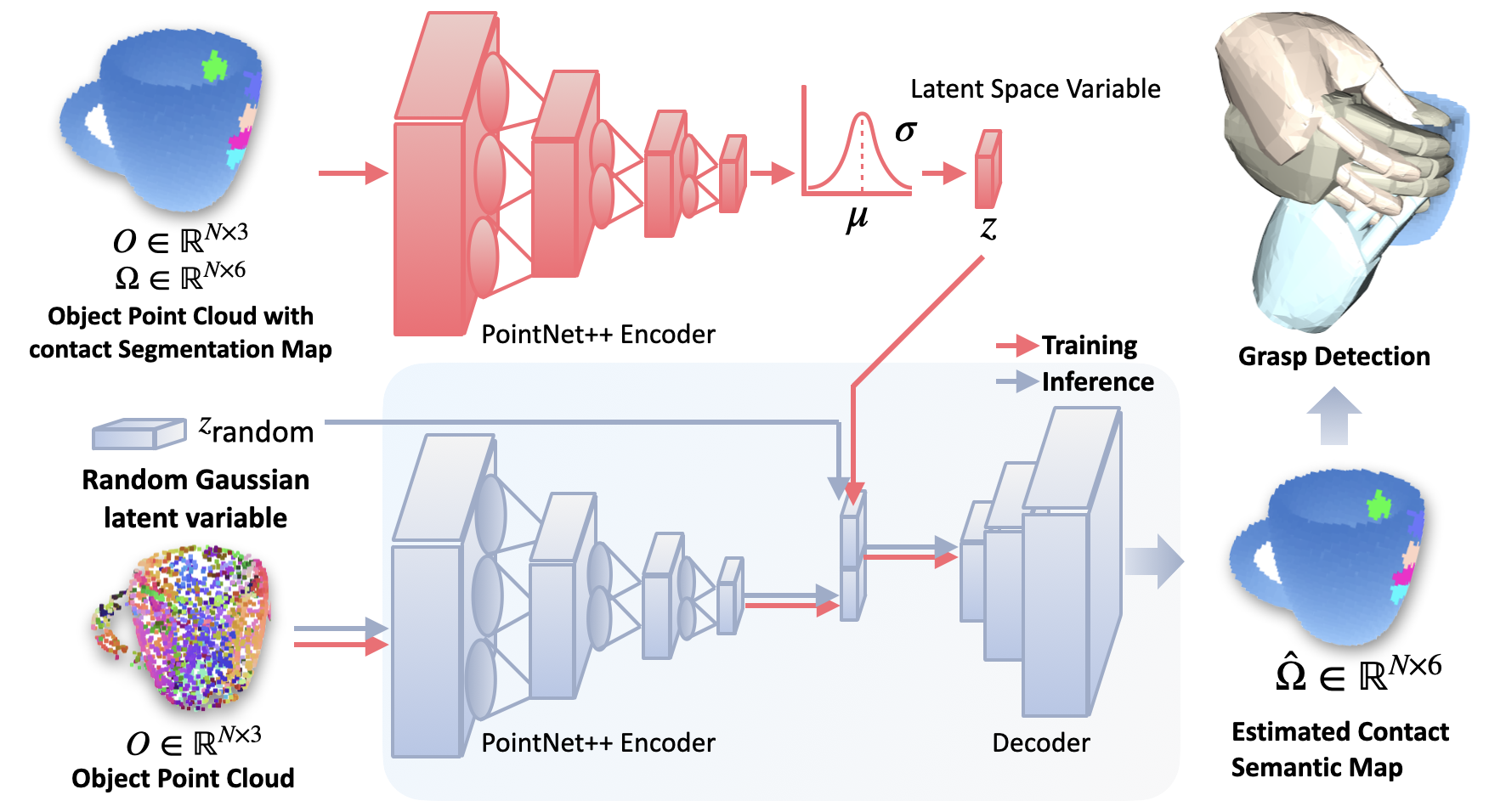}
		\caption{Contact Semantic Map Generation and Grasping Detection.}
		\label{fig.cose_network_pipeline}
	\end{center}
\end{figure}

Given the point cloud data of objects, we employ a novel generative model, Contact Semantic Conditional Variational Autoencoder (CoSe-CVAE), to learn the network for predicting contact semantic maps, as shown in Fig.~\ref{fig.cose_network_pipeline}. In the encoder, the point cloud data $O$ and contact semantic map $\Omega$ are processed through PointNet++~\cite{qi2017pointnet++} to extract both global and local features. The features abstracted from data with contact semantic information are then utilized to predict the mean $\mu$ and variance $\sigma$, from which the latent space variable $z$ is sampled from the data distribution. In the decoder, the latent space variable $z$ and the input point cloud data $O$ are utilized to initially predict the contact semantic maps $\hat{\Omega}$. The encoder and decoder parameters, $\varphi$ and $\theta$, are updated by maximizing the evidence lower bound (ELBO) of log-likelihood of $\log p_{\theta, \varphi}(\Omega \mid O)$, as follows:

\begin{equation}
\begin{aligned}
\centering
    \log p_{\theta, \varphi}(\Omega \mid O) \geqslant  \mathbb{E}_{z \sim Z}&\left[\log p_{\varphi}(\Omega \mid z, O)\right] \\
 -\mathbb{KL}&\left[p_\theta(z \mid \Omega, O) \Vert p_Z(z)\right] \\
 \mathbb{E}_{z \sim Z}\left[\log p_{\varphi}(\Omega \mid z, O)\right]&=\frac{1}{N_o} \sum_{i=0}^{N_o-1}\sum_{c=1}^{C} \omega_{c} \Omega^{c}_{i} \log (\hat{\Omega}^{c}_{i})
\end{aligned}
\end{equation}
where, expectation of ELBO is estimated by weighted cross entropy loss of contact semantic map. $Z$ represents standard normal distribution $\mathcal{N}(0, I)$. $\mathbb{KL}$ denotes the Kullback-Leibler (KL) divergence. $\Omega$ and $\hat{\Omega}$ means the ground truth and estimated contact semantic map within a set of $N_o$ samples and $C$ classes. The class weight is denoted by $\omega$.

\begin{figure*}[htbp]
    \begin{center}
		\includegraphics[width=17cm]{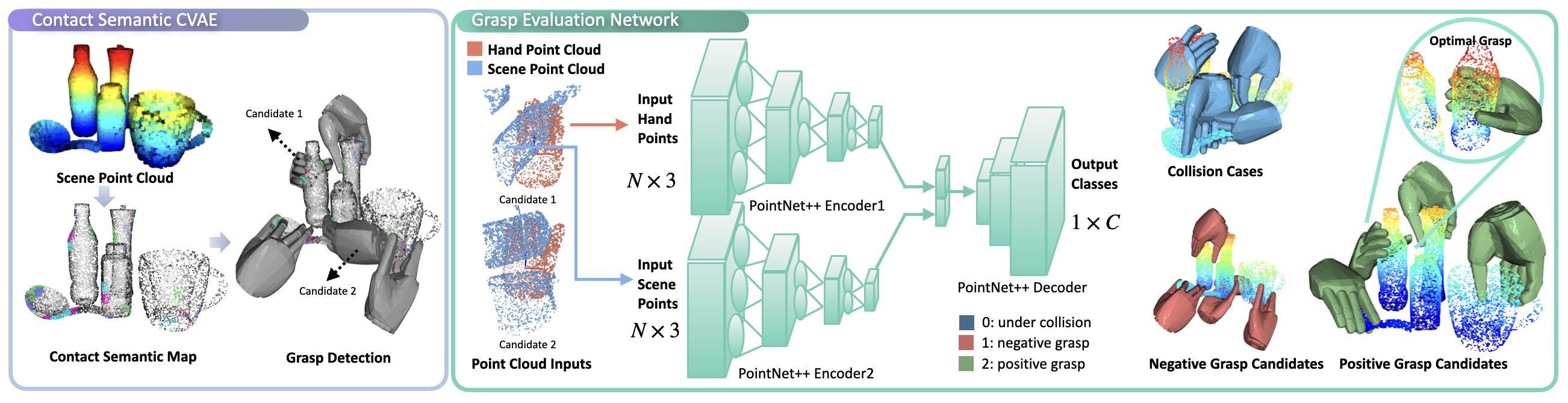}
		\caption{Pipeline of multi-fingered robotic hand grasping network for grasping generation in cluttered environments. Firstly, contact semantic mappings are estimated from object point cloud employing CoSe-CVAE. Secondly, grasp detection method is utilized to generate hand postures from contact prior information. Finally, grasp evaluation network is used to estimated optimal grasp.}
		\label{fig.network_pipeline}
	\end{center}
\end{figure*}

\subsection{Grasping Detection from Contact Semantic Maps}
\label{sec:contact_cvae_and_grasp_detection}

Using the generated contact semantic maps and the surface point clouds of each finger of the robotic hand, we utilize the
correspondence point matching algorithm~\cite{rusinkiewicz2001efficient} to estimate the initial wrist pose. Inspired by GenDexGrasp~\cite{li2023gendexgrasp} and UniGrasp~\cite{shao2020unigrasp}, we optimize the wrist poses and finger tip positions by minimizing energy loss considering proposed contact semantic maps. The joint angles of the manipulator are calculated by differential inverse kinematics library mink~\cite{mink2024}. 
The energy loss function $e$ is formulated as follows:
\begin{equation}
\begin{aligned}
    e =  \sum_{k = 0}^{n - 1} \left|\epsilon_s(v_k, \Omega_k) \right|
\end{aligned}
\end{equation}
where, $\epsilon_s(v_k, O_k)$ represents the signed distance from the fingertip position $v_k$ of the $k$-th robotic finger to the contact points $v_k$, which are labeled with the semantic identifier $k$.

\subsection{Grasp Evaluation Model}
\label{sec:grasp_evaluation_network}

To identify optimal grasp in cluttered environments, we introduce an unified grasp evaluation model PointNetGPD++ to quantify grasp quality, denoted as $q$. As illustrated in Fig.~\ref{fig.network_pipeline}, our network architecture integrates inputs composed of hand surface points $P$ and a partial scene point cloud $H$ surrounding the target grasped object. The partial point cloud is captured relative to the hand’s local frame. The point clouds are processed through two PointNet++ encoders, which extract latent spatial features of the hand and the scene point clouds. These latent features are concatenated and passed through a PointNet++ decoder to predict grasp classification scores. The classification includes three categories: Class 0 (grasp candidates under collision condition), Class 1 (negative grasp candidates), and Class 2 (positive grasp candidates). We employ a multi-class classification loss to guide the network’s training. Specifically, we utilize the categorical cross-entropy loss, defined as:
\begin{equation}
\begin{aligned}
    \mathcal{L} =  -\sum_{x = 0}^{C-1} y_x \log(\hat{y}_x)
\end{aligned}
\end{equation}

where $C$ denotes the number of grasp categories, $y_x$ is the ground truth label for class $x$, and $\hat{y}_x$ represents the predicted probability for class $x$. 

Among positive grasp candidates, the grasp candidate with higher scores are considered optimal.

\setlength{\textfloatsep}{5pt}

\section{Experiment}
\label{sec:experiment}

 \begin{figure}[htbp]
    \begin{center}
		\includegraphics[width=6cm]{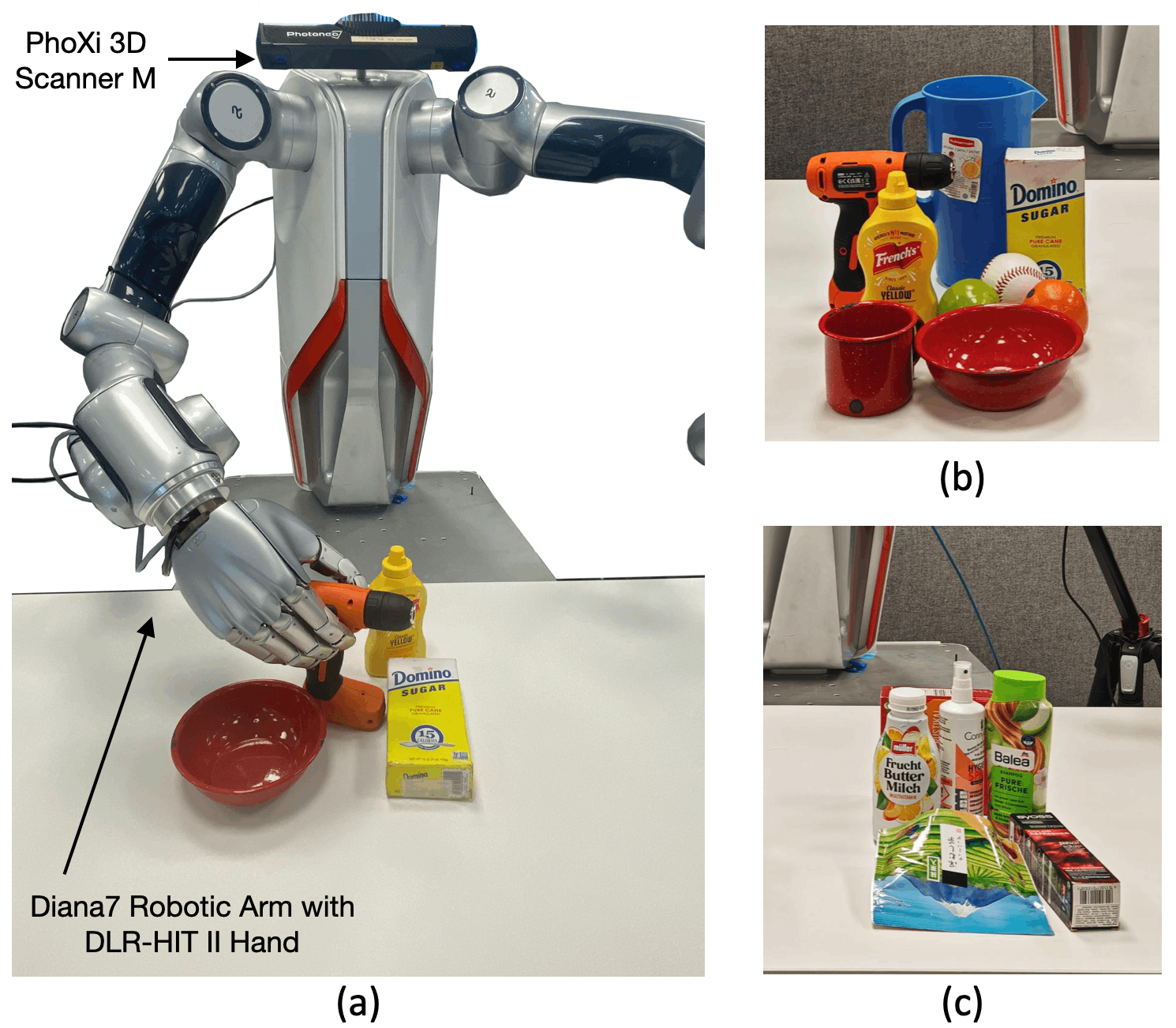}
		\caption{(a) Experiment Setup. (b) Test objects from YCB-Video dataset~\cite{calli2015benchmarking} and (c) Household objects.}
		\label{fig.experiment_setup}
	\end{center}
\end{figure}

\subsection{Experimental Setup}
We establish a platform equipped with Diana7 robot and DLR-HIT II Five-Finger hand~\cite{chen2010experimental} for conducting real-world experiments, as shown in Fig.~\ref{fig.experiment_setup} (a). The control of the robotic hand is achieved through joint impedance control. For capturing the scene's point cloud, we employ PhoXi 3D Scanner M camera. Grasped objects are shown in Fig.~\ref{fig.experiment_setup} (b) and (c).  

During the experiments, the scene point cloud was segmented through instance segmentation and 6D pose estimation to extract object point clouds. For unknown objects, we utilized AnchorFormer~\cite{chen2023anchorformer} for object point cloud completion. After generating the optimal grasp using our pipeline, we employed the SE3 trajectory interpolation algorithm~\cite{RobotCarDatasetIJRR} to plan the robotic arm's trajectory.

\subsection{Multi-fingered Robotic Hand Grasping Dataset}

\begin{figure}[htbp]
    \begin{center}
  \includegraphics[width=8cm]{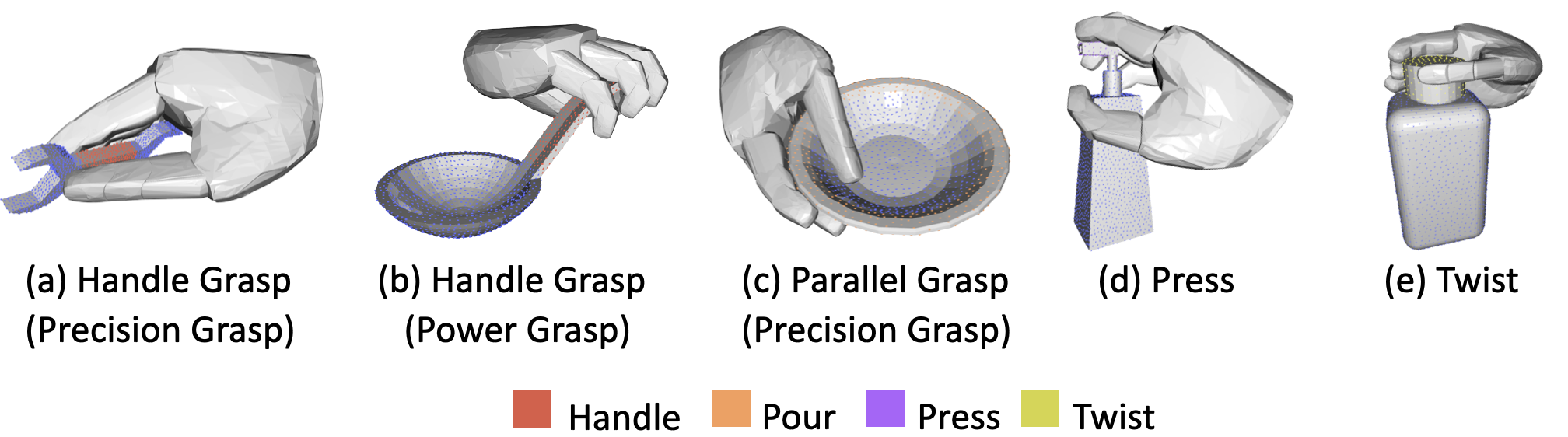}
		\caption{Examples of grasping type considering object affordance maps and different manipulation poses.}
		\label{fig.grasp_type}
	\end{center}
\end{figure}

Our dataset includes 1,521 household models from our previous work~\cite{10011794}, as well as objects from the AffordPose dataset~\cite{jian2023affordpose}. We generated over 2,000 scenes containing cluttered scene point clouds, collision scores, grasp quality, contact semantic maps, and grasp type information, contact distance map. We predict grasp types by voting based on the contact points and the object affordance label from~\cite{jian2023affordpose}, where the grasp type is determined by the affordance with the highest number of votes. The affordance labels include handle grasping, enveloping grasp, pouring, pressing, cutting, and twisting. In this work, we did not explore grasp type information in grasp generation. Dataset examples and grasp types are shown in the Fig.~\ref{fig.pipeline_dataset_generation} and Fig.~\ref{fig.grasp_type}.

\subsection{Model Training}
We train CoSe-CVAE and grasp evaluation models using the cluttered scene point clouds, collision scores, grasp qualities, and contact semantic map data from the dataset. Adam optimizer is utilized with learning rates of 1e-4 and 1e-3 for training models on NVIDIA RTX 6000 Ada GPU. The radius $r$ and height $h$ of the cylinder region used for cropping part of the scene point cloud need to be adjusted according to the gripper dimensions. For DLR-HIT II hand, the radius $r$ and height $h$ are $0.1$ meters and $0.3$ meters, respectively.

\begin{figure*}[htbp]
    \begin{center}
  \includegraphics[width=14cm]{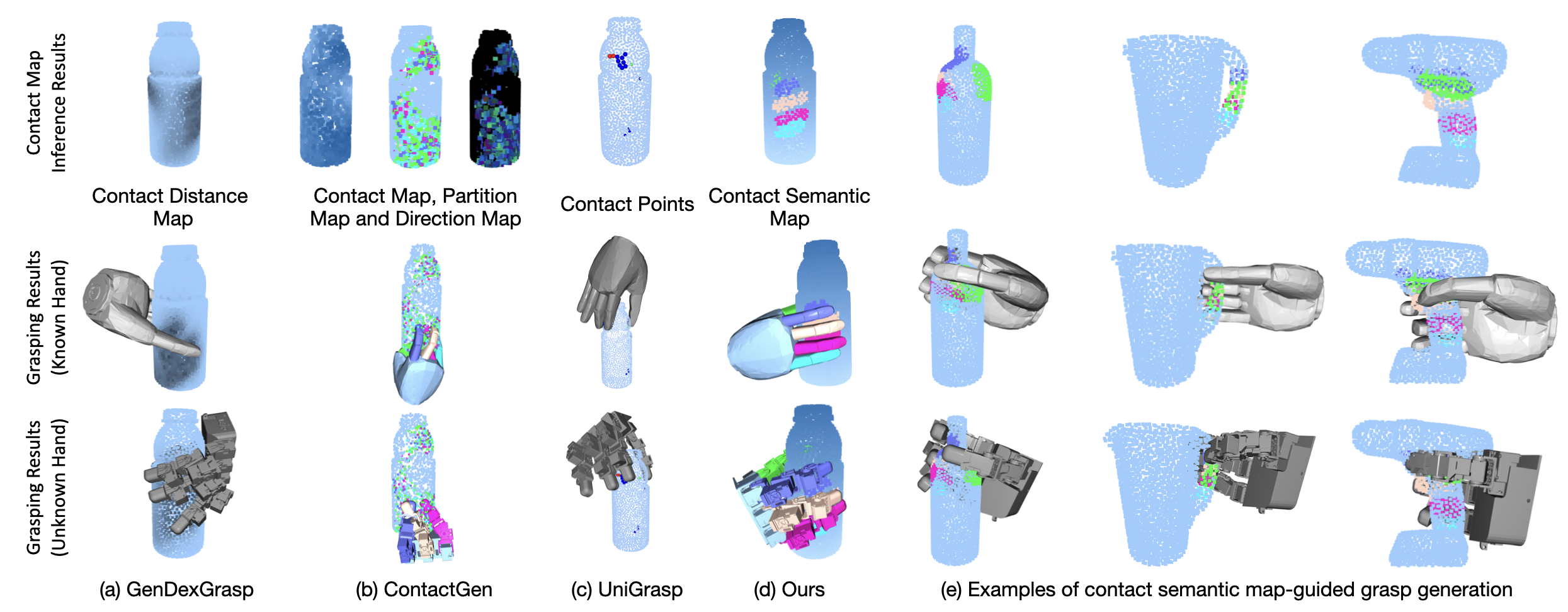}
		\caption{Inference results for grasping with known and unknown hands using SOTA and our methods. (a) Estimated Contact distance map and grasp pose from  GenDexGrasp~\cite{li2023gendexgrasp}. (b) Contact maps and grasp pose from  ContactGen~\cite{liu2023contactgen}. (c) Contact points and grasp candidate based on UniGrasp~\cite{shao2020unigrasp}. (d)-(e) Contact semantic map and grasp candidate from our methods. }
    \label{fig.qualitative_study_vis_single_object}
	\end{center}
\end{figure*}



\subsection{Contact Information-Guided Grasping Generation}
Qualitatively, we compare the grasp generation performance of our method with current contact information-guided grasp generation methods, namely GenDexGrasp~\cite{li2023gendexgrasp}, ContactGen~\cite{liu2023contactgen}, and UniGrasp~\cite{shao2020unigrasp}, as shown in Fig.~\ref{fig.qualitative_study_vis_single_object}.

GenDexGrasp~\cite{li2023gendexgrasp} estimates contact distance map to guide optimization-based grasp generation, but it is prone to sub-optimal solutions in global optimization and suffers from semantic ambiguity. Specifically, the generated grasps do not always align with the predicted contact semantic information, leading to inconsistencies between the grasp poses and the intended contact regions. As shown in Fig.~\ref{fig.qualitative_study_vis_single_object} (a), the generated grasp samples do not match the predicted contact distance map. In contrast, the CoSe-CVAE provides more accurate guidance for grasp generation.

ContactGen~\cite{liu2023contactgen} generates a sequence of contact maps for human grasp synthesis, including a contact map, a part map, and a direction map. However, the complexity of this model leads to cumulative errors across all predicted maps, which propagate to the final grasp result. Moreover, discrepancies in size and joint design between human and robotic hands often prevent multi-fingered from replicating human grasp poses accurately. As a result, ContactGen struggles to generalize well to robotic hands, as illustrated in Fig.~\ref{fig.qualitative_study_vis_single_object} (b). In contrast, our approach utilizes a single contact semantic map, effectively improving stability in multi-finger robotic grasp generation.

UniGrasp~\cite{shao2020unigrasp} predicts contact points one by one, while the CoSe-CVAE utilizes a generative model to predict all contact points simultaneously. As shown in Fig.~\ref{fig.qualitative_study_vis_single_object} (c), the contact points generated by UniGrasp are relatively sparse, and the method exhibits limited diversity in its outcomes. As the number of estimated contact points increases, UniGrasp's ability to account for the relationships between contact points diminishes, resulting in contact point predictions that fail to generate feasible grasp postures. Moreover, representing with sparse contact points introduces additional complexity, as a single grasp sample corresponds to multiple configurations with contact points, increasing the difficulty of model training. CoSe-CVAE is more effective at considering the relationships between contact points, enabling it to generate diverse contact semantic maps. 

Incorporating contact semantic map into the grasping process significantly improves stability of grasp generation, enhances semantic consistency. 


\subsection{Unknown Multi-fingered Robotic Hand Grasp Generation}

To verify the generalization capability of SOTA methods~\cite{li2023gendexgrasp,liu2023contactgen,shao2020unigrasp} and our CoSe-CVAE across different robotic hands, we estimate grasp candidates using known hand, DLR-HIT II hand, and unknown robotic hand, four-fingered LEAP hand~\cite{shaw2023leap} from identical contact maps. The results are shown in Fig.~\ref{fig.qualitative_study_vis_single_object} (d) and (e). 
Compared to SOTA methods, CoSe-CVAE provides more stable and precise guidance for grasp pose estimation of unknown robotic hands.

\subsection{Grasping from Cluttered Scenes}
\begin{figure}[htbp]
    \begin{center}
  \includegraphics[width=8cm]{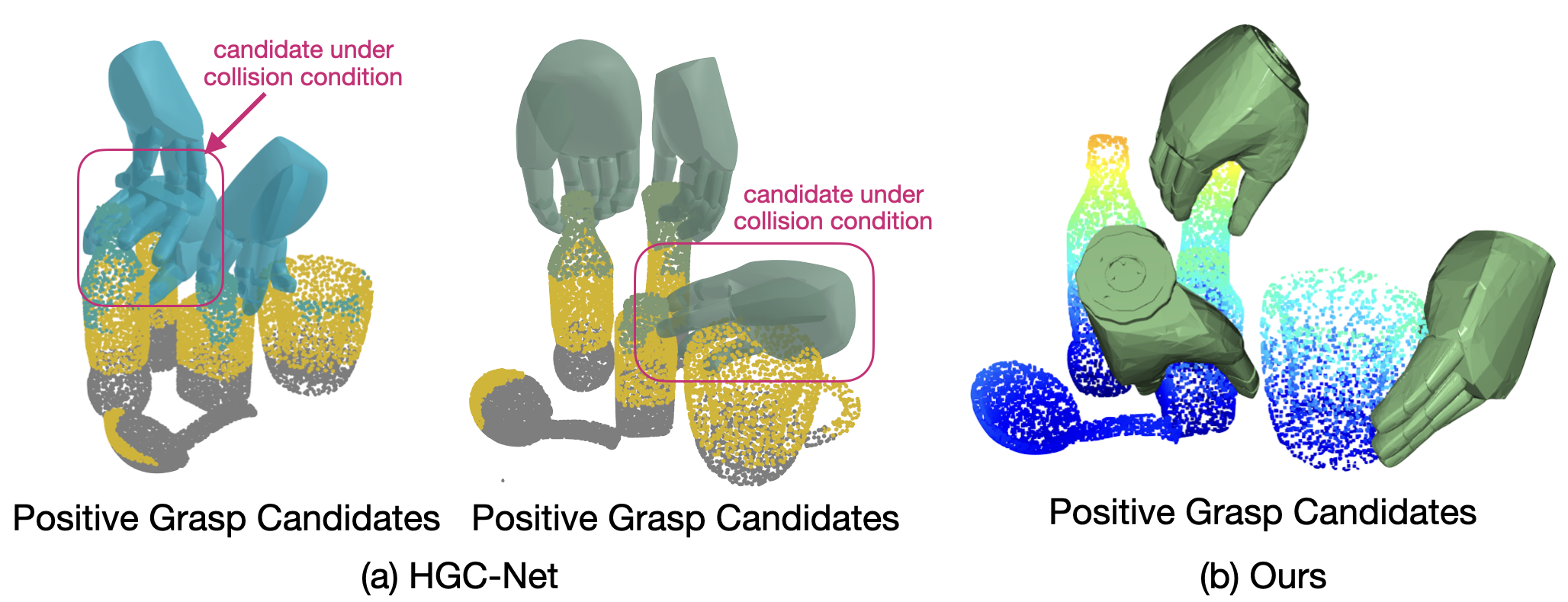}
		\caption{Inference results for cluttered scenes. (a) Grasp candidates based on HGC-Net~\cite{li2022hgc}. (b) Positive grasp candidates based on our methods. }
		\label{fig.qualitative_study_vis_cluttered}
	\end{center}
\end{figure}

To evaluate the performance in cluttered scenarios, we perform comparison experiments using our method and the SOTA approach, HGC-Net~\cite{li2022hgc}. The planning results for cluttered environments with unknown objects are presented in Fig.~\ref{fig.qualitative_study_vis_cluttered}. The optimal grasps generated by our model consistently outperform those of HGC-Net, delivering higher-quality results. 
In cluttered real-world scenarios, HGC-Net often predicts grasp poses that result in unintended collisions with objects. This premature finger contact can lead to grasp failures, reducing overall grasp success rates. By incorporating a contact semantic map between the perception and grasping processes, our pipeline is capable of assessing grasps in cluttered environments, consistently identifying the optimal grasp.

\subsection{Quantitative Grasping Experiments}
We conduct comparison experiments in grasping a single object using known and unknown hands, as well as grasping from cluttered environments using a known hand. For each setting, we perform $150$ grasp attempts per method. A grasp is considered successful if the robotic hand securely lifts the object and maintains a stable grip for at least two seconds. Any grasp predicted as successful, but resulting in a collision, will be automatically considered a failure. The success rate is determined by calculating the proportion of successful grasps out of the total number of attempts. The quantitative results are shown in Tab.~\ref{tab.grasp_comparison_combined}.

\begin{table*}[htbp]
\centering
    \vspace*{-2mm}
	\caption{Quantitative results of real-world grasping experiments}
	\label{tab.grasp_comparison_combined}
	\vspace*{-3mm}
\begin{threeparttable}
\begin{tabular}{ccccccccc}  %
\toprule
 \multirow{6}{*}{\makecell*[l]{\textbf{Method}}}&
\multicolumn{6}{c}{ \multirow{1}*{Success Rate(\%)}\rule{0pt}{0.2cm}}\\
\\[-0.7em]
\cline{2-7}
&
\multicolumn{2}{c}{ \multirow{1}*{Single-Object Scene}\rule{0pt}{0.3cm}}&
\multicolumn{2}{c}{ \multirow{1}*{Single-Object Scene}\rule{0pt}{0.3cm}}&
\multicolumn{2}{c}{ \multirow{1}*{Cluttered Scene}\rule{0pt}{0.3cm}}\\
\\[-0.7em]
\cline{2-7}
&
\multicolumn{2}{c}{ \multirow{1}*{Known Hand}\rule{0pt}{0.3cm}}&
\multicolumn{2}{c}{ \multirow{1}*{Unknown Hand}\rule{0pt}{0.3cm}}&
\multicolumn{2}{c}{ \multirow{1}*{Known Hand}\rule{0pt}{0.3cm}}\\
\\[-0.7em]
\cline{2-7}
& \makecell*[c]{Household} & \makecell*[c]{YCB} & \makecell*[c]{Household} & \makecell*[c]{YCB} & \makecell*[c]{Household} & \makecell*[c]{YCB}\\
\midrule
GenDexGrasp~\cite{li2023gendexgrasp} & 62.7 & 58.7 & 60.7 & 55.3 & - & - \\
\\[-0.7em]
ContactGen~\cite{liu2023contactgen} & 64.7 & 63.3 & 62.0 & 59.3 & - & - \\
\\[-0.7em]
UniGrasp~\cite{shao2020unigrasp} & 71.3 & 70.7 & 70.7 & 64.7 & - & - \\
\\[-0.7em]
HGC~\cite{li2022hgc} & 78.7 &  74.0 & - & - & 70.7 & 70.0 \\
\\[-0.7em]
\rowcolor{light-blue}

\textbf{Ours} & \textbf{85.3} &  \textbf{76.7} & \textbf{80.0} &\textbf{73.3} & \textbf{78.0} &\textbf{75.3}   \\
\bottomrule

\end{tabular}
    \vspace*{0mm}

    \end{threeparttable}
    
    \vspace*{-2mm}
\end{table*}
\setlength{\textfloatsep}{5pt}

\subsubsection{Single-Object Grasping using Known Hand}
Using our method, the average grasping success rate reaches 81.0\% in single-object scenes, surpassing other baseline approaches~\cite{li2023gendexgrasp,shao2020unigrasp,li2022hgc,liu2023contactgen}. Our grasp detection methods improve the accuracy of grasping candidate selection. 
\subsubsection{Contact Information-Guided Grasping using Unknown Hand}
We conduct comparison experiments of real-world robotic grasping using the unknown hand. 
The average success rate of our pipeline reaches 76.7\%, outperforming the SOTA methods~\cite{li2023gendexgrasp,liu2023contactgen,shao2020unigrasp}. We also conduct real-world grasping experiments from cluttered experiments using the unknown hand based on our approach. The average success rate achieves 74.0\%.

\subsubsection{Grasping from Cluttered Scenes} 
In grasping experiments from cluttered scenes, our method achieves a 76.7\% average success rate. Our evaluation model substantially aids in filtering out unfeasible grasps for cluttered scenes. 

\begin{figure}[htbp]
    \begin{center}
  \includegraphics[width=7cm]{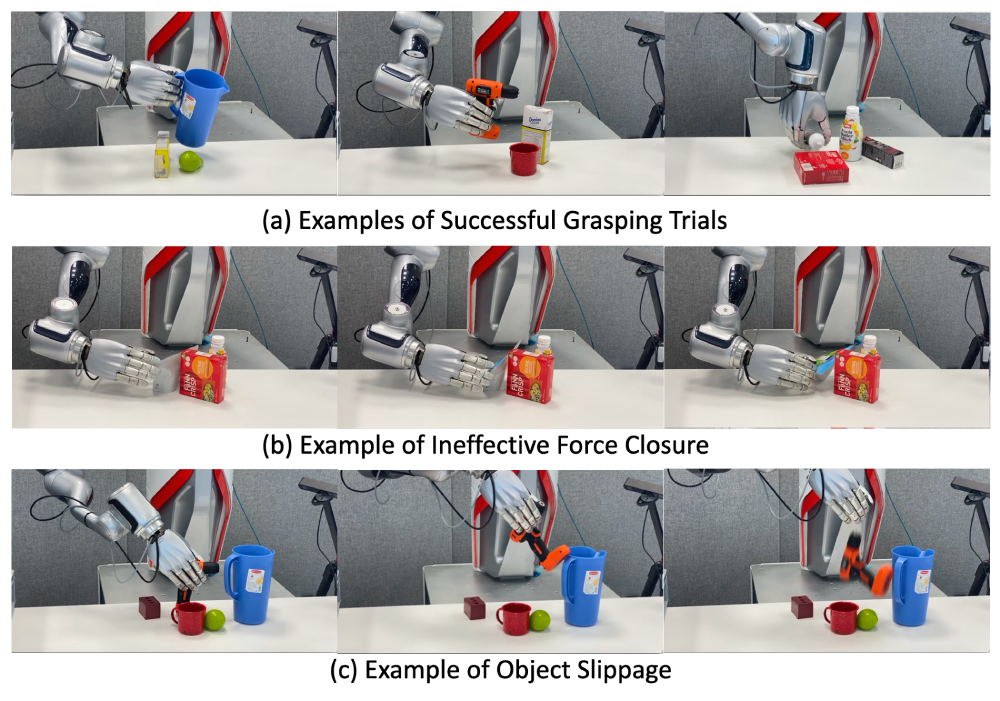}
		\caption{Examples of successful and failed grasping trials.}
		\label{fig.example_of_grasp_trials}
	\end{center}
\end{figure}

\subsection{Grasping Failure Analysis}
Most grasping failures are due to collisions with the surrounding environment, object slippage and ineffective force closure, as shown in Fig.~\ref{fig.example_of_grasp_trials}. Our method results in fewer grasping failures due to collisions with the surrounding environment compared to HGC-Net~\cite{li2022hgc}. In grasping experiments from cluttered scenes, our collision failure rate is 4.0\%, whereas HGC-Net's is 10.7\%, representing a 62.5\% reduction in collision failures. 
Errors in joint angles during grasp generation cause slippage, while inaccuracies during the process prevent proper force closure. 
This issue can be mitigated through tactile-based manipulation, which we will explore in future work.

\section{Limitations and Future Work}
The current dataset has a much higher number of pinch grasps compared to other grasp types, as the data generation algorithm~\cite{wang2023dexgraspnet} cannot explicitly generate candidates with specific grasp types. 
Future work will incorporate contact information for grasp type-aware generation. 
Secondly, our focus is on grasp pose generation, with robot trajectories in experiments determined via path planning. We leave policy-based trajectory generation as a direction for future work.

\section{Conclusions}
\label{sec:conculsion}

We propose a novel semantic contact information-guided grasp generation method for multi-fingered robotic hands in single-object and cluttered environments. First, the CoSe-CVAE model predicts diverse contact semantic maps between the hand and the object from the object’s point cloud. The grasp detection method then estimates grasp poses based on these semantic contact maps. Furthermore, our proposed grasp evaluation network PointNetGPD++ utilizes both scene and robotic hand point clouds to predict grasp quality, selecting the optimal grasp in cluttered scenes. 

Qualitative real-world experiments demonstrate that our CoSe-CVAE model can reliably generate hand-object contact semantic information, significantly enhancing the stability of grasp generation based on contact information using known and unknown hands, outperforming SOTA methods~\cite{li2023gendexgrasp,shao2020unigrasp,liu2023contactgen}. By incorporating the geometric characteristics of the robotic hand, the proposed grasp evaluation model can more effectively assess the grasp quality of multi-fingered hands in single-object and cluttered environments, outperforming SOTA methods~\cite{li2022hgc,liu2023contactgen,li2023gendexgrasp,shao2020unigrasp}. Quantitative comparisons in real-world experiments further show that our method achieves a higher grasp success rate than these SOTA methods, with an average success rate of 81.0\% in single-object scenarios and 76.7\% in multi-object scenarios. 
Additionally, average success rate of grasp experiments using unknown robotic hand reaches 76.7\% in single-object scenes, surpassing SOTA methods~\cite{li2023gendexgrasp,shao2020unigrasp,liu2023contactgen} by at least 9.0\%. 


%

\bibliographystyle{IEEEtran}
\bibliography{main}

\end{document}